\documentclass[letterpaper, 10 pt, conference]{ieeeconf}  
\usepackage[
    left=48pt,        
    right=48pt,       
    top=57pt,         
    bottom=46pt,      
]{geometry}
\usepackage{xcolor}
\makeatletter
\let\NAT@parse\undefined
\makeatother
\usepackage[colorlinks,linkcolor=blue,citecolor=green]{hyperref}
\usepackage{amsmath,amssymb,amsfonts}
\usepackage{algpseudocode}
\usepackage{graphicx}
\usepackage{textcomp}
\usepackage{wrapfig}
\usepackage[numbers,sort&compress]{natbib}
\usepackage{amsmath}
\usepackage{float}
\usepackage{bm}

\usepackage{overpic}
\usepackage[justification = centering,labelsep = period]{caption}
\usepackage{algorithm}
\usepackage{array}
\usepackage[caption=false,font=normalsize,labelfont=sf,textfont=sf]{subfig}
\usepackage{stfloats}
\usepackage{url}
\usepackage{verbatim}
\usepackage{caption} 
\usepackage{color}
\usepackage{booktabs}
\usepackage{multirow}
\usepackage{makecell}
\usepackage{tabularx}
\usepackage{cleveref}
\crefname{figure}{Fig.}{Figs}  
\Crefname{figure}{Figure}{Figures}  
\crefname{table}{Table}{Tables}  
\Crefname{table}{Table}{Tables}  
\usepackage[numbers,sort&compress]{natbib}

\floatname{algorithm}{{Algorithm}}

\IEEEoverridecommandlockouts 
\title{\LARGE \bf
A Novel Method with Encoder-Decoder for Cross-Sensor Adaptation in Surface Shape Sensing with Sparse Strain Sensors}
\author{Shuo Wang, Heng Luo,~\IEEEmembership{Student Member,~IEEE}, Dian Jin and Xiaoming Tao,~\IEEEmembership{Member,~IEEE}} 

\begin{document}
\newgeometry{top=60pt, left=48pt, right=48pt, bottom=46pt}
\maketitle

\begin{abstract}
\textcolor{black}{Performance variations in sensor arrays, caused by intrinsic differences or installation conditions, can lead to inconsistent results during shape sensing. To obtain accurate results, a large amount of data is usually required, and a separate model must be retrained for each sensor array, thereby increasing the cost and time of data acquisition, transmission, and computation. To address this issue, this work} proposes an encoder-decoder architecture for surface shape sensing based on sparse strain sensors \textcolor{black}{and further} incorporates meta-learning and few-shot adaptation strategies \textcolor{black}{to enable adaptation across different groups of sensor arrays}. Experimental results demonstrate that, after the cross-sensor adaptation, a newly deployed sensor array achieves a sensing error \textcolor{black}{of approximately} 4.0 mm \textcolor{black}{relying on less than 5.0\% newly labeled data and requiring an adaptation time of under 1 second, which represents a substantial improvement from 23.0 mm error without adaptation and 20-minute data collection time required to train a new model. Moreover, the number of points with errors below 5.0 mm increased by more than 65.0\%. These results indicate that the proposed method can substantially reduce the cost and training burden of surface shape sensing, and it has broad potential applications in soft robotics and wearable devices.}
\end{abstract}


\section{Introduction}
In recent years, soft and flexible structures have been increasingly incorporated in robotic systems to be used for surgery \cite{lee2017nonparametric} or rehabilitation \cite{polygerinos2015soft}. The actual shape and posture of the robot are inevitably influenced by their deformability with the interactive environment \cite{laschi2016soft}. Flexible electronic patch-like sensors have been developed for monitoring physiological signals or other functionalities. \textcolor{black}{Due to the complex and dynamic nature of the human body, these patches are susceptible to deformation-induced errors when they are conformally attached to anatomical sites. For example, the performance of flexible ultrasound transducers  \cite{chen2023flexible,1} is influenced by the deformation of the sensor attached to the surface of the human body. Similar challenges arise in soft robotics: when a robot interacts with a soft object, accurate shape sensing becomes difficult, and force measurements alone often provide insufficient guidance.}

\textcolor{black}{Consequently, directly quantifying surface morphology has emerged as a key research topic.} Non-contact techniques \textcolor{black}{using external device include optical marker-based methods \cite{4,dobrzynski2011contactless}, stereo camera and laser imaging, detection and ranging-based approaches \cite{sansoni2009state}, electromagnetic tracking systems \cite{5} and others.} However, they all require complex and high-precision external equipment, and then derive the shape deformation from the measured surface feature data. The necessity of additional sophisticated equipment limits their flexibility, rendering them bulky, difficult to integrate, and susceptible to environmental noise interference, which implies great obstacles to their adaptation in real-time robotic systems or other soft wearable systems.

\textcolor{black}{Employing self-contained sensors to capture local deformation and orientation data constitutes an alternative method. Polymeric optical fiber was used for one-dimensional bending shape detecting in a small specific area as wearable sensors \cite{wang2021low}. Electrical impedance tomography} shows the possibility of using electrical impedance or mutual inductance of shape characteristics \cite{7,16,17}. Microelectromechanical systems enable the incorporation of a lot of miniaturized sensors that provide localized data, facilitating the complete reconstruction of global surface geometry \cite{hermanis2015acceleration}. Fiber Bragg gratings measure point-distributed strain and achieve high-precision shape sensing \cite{2,8,11,wang2021large,dong2024robotic}. \textcolor{black}{Low-cost  strain sensors or gauges can also be used for surface shape sensing \cite{9,10,12,liu2025model}, which has more flexibility and less complicated. Compared to optical ones, the electronic systems do not require light sources and bulky spectral equipment}, \textcolor{black}{which makes them easier to integrate and deploy} in portable or volume-constrained applications while still achieving high sensing accuracy.

Strain sensors have been \textcolor{black}{shown to enable shape mapping by using deep learning methods or geometric formulations \cite{9,10,12,liu2025model}. However, the performance characteristics can vary substantially \textcolor{black}{due to individual variations, particularly for lab-made or low-cost sensors.} Even commercial sensors have significant variations between batches, which has a great influence on the final results \cite{HBM2021strain}. When such sensors are deployed or applied on a large scale, current deep learning methods typically require continuous acquisition of a large amount of labeled data for model training, which inevitably consumes substantial time and computational resources and increases the overall system cost.}

\textcolor{black}{In practical applications, performance degradation caused by intrinsic differences among arrays or by installation misalignments can be regarded as a problem of cross‑sensor domain adaptation or transfer learning. Among various sensing applications, one solution to address these issues is to map data from different sensors into a common latent space and to align their representations \cite{luo2025cross}. Few-shot unsupervised domain adaptation is successfully utilized to reduce cross-sensor domain gaps with tactile sensors \cite{jing2025reducing}. For piezoresistive sensor array, the local message passing network with message passing mechanism enables single touch data to learn correlations between neighboring sensor subelements, which can narrow the domain shift of the single touch and multi touch datasets \cite{kim2021single}}. \textcolor{black}{Many other sensors, such as magnetic sensors, gas sensors, and inertial measurement units, have been calibrated using few-shot learning to reduce discrepancies between sensor measurements and theoretical models \cite{20}, while meta-learning approaches have been employed to further improve their measurement stability and overall performance \cite{tritschler2024meta,tianliang2023development}.} \textcolor{black}{When different sensors are used, differences among sensors and noise from various sources can degrade or even invalidate existing models, underscoring the practical importance of cross‑sensor adaptation. It is therefore crucial to account for variations arising from sensor characteristics or other factors, as these can shift the feature distribution and lead to significant errors in shape sensing when using strain sensors.}

\textcolor{black}{To address this challenge, this study adopts two deep learning methods: encoder–decoder neural network and meta-learning. The encoder–decoder network is a supervised architecture in which an encoder compresses the input into a latent representation and a decoder reconstructs the target output, enabling end-to-end modeling of complex nonlinear relationships between high-dimensional inputs and outputs \cite{encoder-decoder}. Meta-learning trains models across multiple tasks so that they can rapidly adapt to new tasks or domains with only a small amount of data, thereby improving sample efficiency, adaptation speed, and robustness to distribution shifts \cite{tritschler2024meta,tianliang2023development}.}

\textcolor{black}{The encoder–decoder neural network is used to map strain sensor readings to relative feature point coordinates obtained from a stereo camera, and is combined with meta-learning and few-shot adaptation to achieve efficient cross-sensor adaptation. A new group of strain sensor array can thus be calibrated with only a small amount of data under diverse deformations, enabling rapid and accurate surface shape sensing. Consequently, the strain sensor array can subsequently operate independently of the camera in practical use, and multiple arrays performing the same sensing task can achieve high precision without extensive additional data collection or repeated model retraining. Thus, the system and model can easily be tailored or integrated to specific applications. The main contributions of this study are:}

1) \textcolor{black}{A new platform with a simple backend acquisition device and computational model for surface shape sensing with sparse strain sensor data that can solely be used for calculation after model training for shape sensing.}

2) \textcolor{black}{An encoder–decoder framework with a Transformer encoder and a graph neural network decoder, incorporating the graph structure of the measured surface into the loss function and using virtual tasks constructed from strain sensor data in meta‑learning to enable adaptation to new groups of sensor arrays with only a few labeled samples.}

3) \textcolor{black}{Experiments with multiple sensor arrays, demonstrating high sensing accuracy with less than $5.0\%$ new labeled data and enabling fast adaptation ($\leq 1 s$), thereby greatly reducing data collection and model training costs for new groups of sensor arrays. }

\section{Methods}
 
\subsection{Mapping of Strain Values to Coordinates}
\textcolor{black}{The analysis is conducted for a small-scale, thin surface whose thickness is much smaller than its in-plane dimensions. According to elasticity theory, in this type of surface, local deformation generates a smooth global strain field, through which strain information propagates across the material. As a result, strain measurements at a limited number of locations are, in principle, sufficient to reconstruct the surface deformation with acceptable accuracy.} 

\begin{figure}[h]
\centering
\vspace{0cm}
\includegraphics[width=0.28\textwidth]{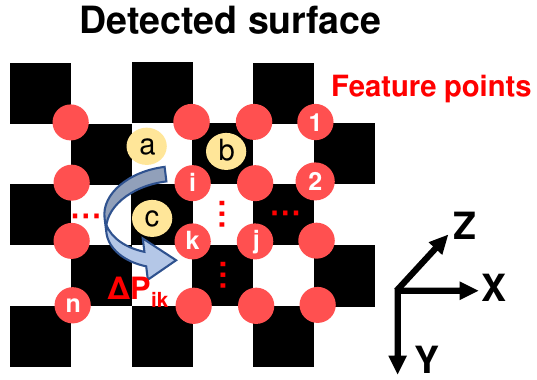}
\caption{\textcolor{black}{\small Mapping relationship.}}
\label{Corners}
\vspace{-0.7cm}
\end{figure}

\textcolor{black}{To enable shape sensing, a checkerboard pattern with $n$ red feature points is attached to the surface, as illustrated in \cref{Corners}. Point $k$ is chosen as the coordinate origin, and the deformation is described by the relative displacement of each feature point with respect to point $k$. All coordinate changes are thus expressed in this local reference frame.}

\textcolor{black}{To simplify the analysis, we focus on a feature point $i$ and consider three neighboring regions, labeled $a$, $b$ and $c$. When region $a$ deforms, the corresponding strain $\epsilon_{a}$ is measured, and the strain information is assumed to influence point $i$ through a latent function $g_{a}$. Similarly, strains $\epsilon_{b}$ and $\epsilon_{c}$ in regions $b$ and $c$ affect point $i$ through $g_{b}$ and $g_{c}$, respectively. The combined effect of these transmitted strain contributions results in the observed displacement of point $i$, which is described by a mapping function $f$ acting on the latent responses. Direct determination of $g_{a}$, $g_{b}$, $g_{c}$ and $f$ is highly challenging and generally intractable. Instead, these intermediate functions are implicitly eliminated, and an effective mapping $F_{i}$ is introduced that directly relates the displacement of point $i$ to the measurable strains in regions $a$, $b$ and $c$. For any other point on the measured surface, if several strain measurements in its surrounding region are available, an analogous relationship can be derived, and this strain information can be further propagated to other points. Consequently, if $m$ strain sensors are distributed on the surface to obtain strain measurements, the coordinate change of the $i_{th}$ point can be expressed as \eqref{deqn_ex1}:
}

\textcolor{black}{
\begin{equation}
\label{deqn_ex1}
\Delta{P_{ik}}^{\mathsf{T}} =  F_{i\_{direct}}(\epsilon_{s_1}, \epsilon_{s_2}, \ldots \epsilon_{s_m})\quad i = 1,2 \ldots n
\end{equation}}

\textcolor{black}{
where $\Delta{P_{ik}}\in\mathbb{R}^3$, it represents the coordinate change of point $i$ in $X$, $Y$ and $Z$ axis.
$F_{i\_direct}$ means the direct mapping from strain measurements to the coordinates' change of the $i_{th}$ point on the surface. $\epsilon_{s_m}$ means the measurement of the $m_{th}$ sensor, and $n$ represents the number of points. Therefore, the objective of the surface shape sensing problem can be defined as establishing a series of mapping relationships between the readings of the strain sensor array (as the input) and the changes in the coordinates of feature points on the checkerboard (as the output) during deformation.}

\subsection{Proposed Method}
\textit{1) Ground truth computing with a stereo camera:} Sub-pixel corner detection is utilized to detect the coordinate changes of feature points when the surface has deformation. The target is a checkerboard test paper attached on a soft polyvinyl chloride (PVC) surface, and the world coordinates of the feature points are obtained via a stereo camera. The sub-pixel corner detection and stereo camera calculation methods can be learned from \cite{22,23}, which are commonly employed techniques in the realm of computer vision.

\textit{2) Model:}
\label{model_parameters}
The proposed model comprises two main modules: an encoder with sensor embeddings and \textcolor{black}{Transformer} encoder blocks, and a decoder module with graph attention convolution modules. A detailed description of each module is shown below.

\textit{\quad 2.1) Encoder:}
In the encoder module, a linear embedding from 6 to 128 dimensions is utilized to augment the sparse strain data into a richer representation, effectively simulating a virtual dense strain sensing field. Subsequently, upon obtaining an increased number of channels, two transformer encoder blocks with four attention heads perform feature extraction, capturing latent relationships in the embedded strain representation.

\textit{\quad 2.2) Decoder:}
In order to maintain a lightweight model design, the decoder module contains a node embedding and residual connection parts, followed by two graph attention convolution layers (GATconv), then the output will be projected to the three-dimensional tensor which corresponds to $(x, y, z)$ of points. The embedding layer contains learnable parameters of shape $(20,128)$, followed by two GATconv layers with dimensions $(128,128)$ and $(256,128)$, using 2 and 1 attention heads, respectively.

\textit{\quad 2.3) Graph construction and curvature representation:}
\label{ce}
 \textcolor{black}{A $4 \times 5$ grid with 20 nodes} corresponds to the target surface. Each node and its edges are recorded and used to construct the edge indices of the grid. 
\textcolor{black}{Taking the $X$ direction as an example, the discrete second derivative at row $i$ and column $j$ is approximated as:
\begin{equation}
D_{x(i,j)} = \frac{d_{x(i+1,j)} - 2\,d_{x(i,j)} + d_{x(i-1,j)}}{h^2}
\end{equation}
where $h$ is the constant spacing between adjacent grid rows and columns, and $d_{x(i,j)}$ denotes the displacement of point at row $i$ and column $j$.
To obtain a curvature-related quantitative descriptor of bending that has the same physical dimension as the displacement, we define:
\begin{equation}
\label{def}
C_{x(i,j)} = D_{x(i,j)} \, h^2 
\end{equation}
The corresponding quantity in the $Y$ direction, $C_{y(i,j)}$, is defined analogously.
We then define $C_{\mathrm{overall}}$ as:
\begin{equation}
\label{curvature}
C_{\mathrm{overall}} = \mathrm{mean}\big(|C_{x(i,j)}|\big) + \mathrm{mean}\big(|C_{y(i,j)}|\big),
\end{equation}
In \eqref{curvature}, $i = 2,3$ and $j = 2,3,4$. For boundary nodes ($i = 1,4$ or $j = 1,5$), $|C_{x(i,j)}|$ or $|C_{y(i,j)}|$ is set to zero because the finite-difference stencil is not available. $|\cdot|$ denotes the Euclidean norm, and
$C_{\mathrm{overall}}$ is proportional to the overall curvature and has the same physical dimension as the displacement.}
\textit{3) Loss function:}
The loss function of the model \textcolor{black}{consists of two main components: the mean squared error and a curvature term $C_{overall}$, denoted by $\mathcal{L}_{\mathrm{curvature}}$ and weighted by $\lambda$ in this paper. Incorporating the $\mathcal{L}_{\mathrm{curvature}}$ as a loss term helps suppress excessively abrupt coordinate changes during prediction, ensuring that the model preserves the smooth physical properties of the surface rather than focusing solely on coordinate accuracy.} The complete loss function can be represented as \eqref{loss_func}:
\begin{equation}
\label{loss_func}
\mathcal{L} = \mathcal{L}_{\mathrm{MSE}} + \lambda \cdot \mathcal{L}_{\mathrm{curvature}}
\end{equation}
where $\mathcal{L}_{\mathrm{MSE}}$ denotes the MSE loss, and $\lambda$ is a tunable hyperparameter set to 0.01 in this study.

\textit{4) Meta-learning:}
We aim to leverage meta-learning to enable rapid adaptation of the model across different sensor arrays.

\begin{figure}[h]
\centering
\vspace{0.15cm}
\includegraphics[width=2.9in]{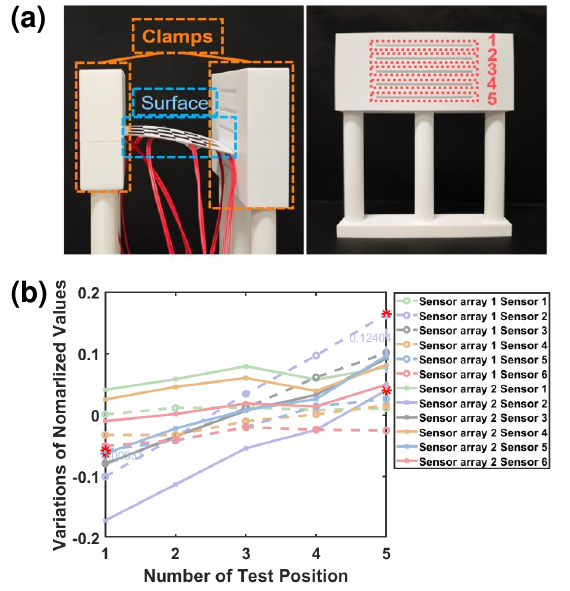}
\caption{\small Performance of two groups of strain sensors. (a) Five level of bending test platform. (b) Results of each strain sensor. }
\label{test position}
\vspace{-0.3cm}
\end{figure}

\textit{\quad 4.1) \textcolor{black}{Generation of virtual tasks:}}
To avoid the high cost of collecting large-scale data from diverse tasks, we generated virtual tasks for meta-learning based on characteristic differences among several strain sensors. \cref{test position} (a) visualizes the bending platform of a PVC with \textcolor{black}{a} strain sensor array attached, and tests can be conducted in five different levels of bending. Then, the normalized variations of two groups of strain sensors can be seen in \cref{test position} (b). The specific arrangement of the sensor array can be seen in Section \ref{sec:Experiments}. According to the tests, the responses of different sensor arrays to the same deformation exhibit variability and fluctuations. \textcolor{black}{Therefore, during the generation of random tasks, data augmentation is applied to the input data.} The augmentation includes scaling, offset, correlation noise among sensors, and scaling of the noise level. \textcolor{black}{In this study, the scaling factor is sampled from 0.8 to 1.2, the random offset from 0 to 0.4, and the correlation noise from 0.7 to 1.0. In addition, an extra scaling factor for the correlation noise level is applied, with values between 0.05 and 0.15.} The outputs corresponding to the initial source input data used for virtual task generation are employed as labels, serving as the outputs of the virtual tasks. Finally, these inputs and the corresponding outputs are used to train as virtual tasks. 

\textit{\quad 4.2) Training and model adaptation process:}
For meta-learning, it involves parameter updates in two stages: inner-loop updates and outer-loop updates. For model adaptation to new groups of strain sensors, only the decoder is fine-tuned for several steps. The adaptation steps, learning rate, unfreeze parameters, \textcolor{black}{and adaptation ratio can all be adjusted to fine-tune the model. The meta-learning training procedure is shown in Algorithm~\ref{algorithm}.}

\begin{figure}[h]
\centering
\vspace{0.15cm}
\includegraphics[width=2.4in]{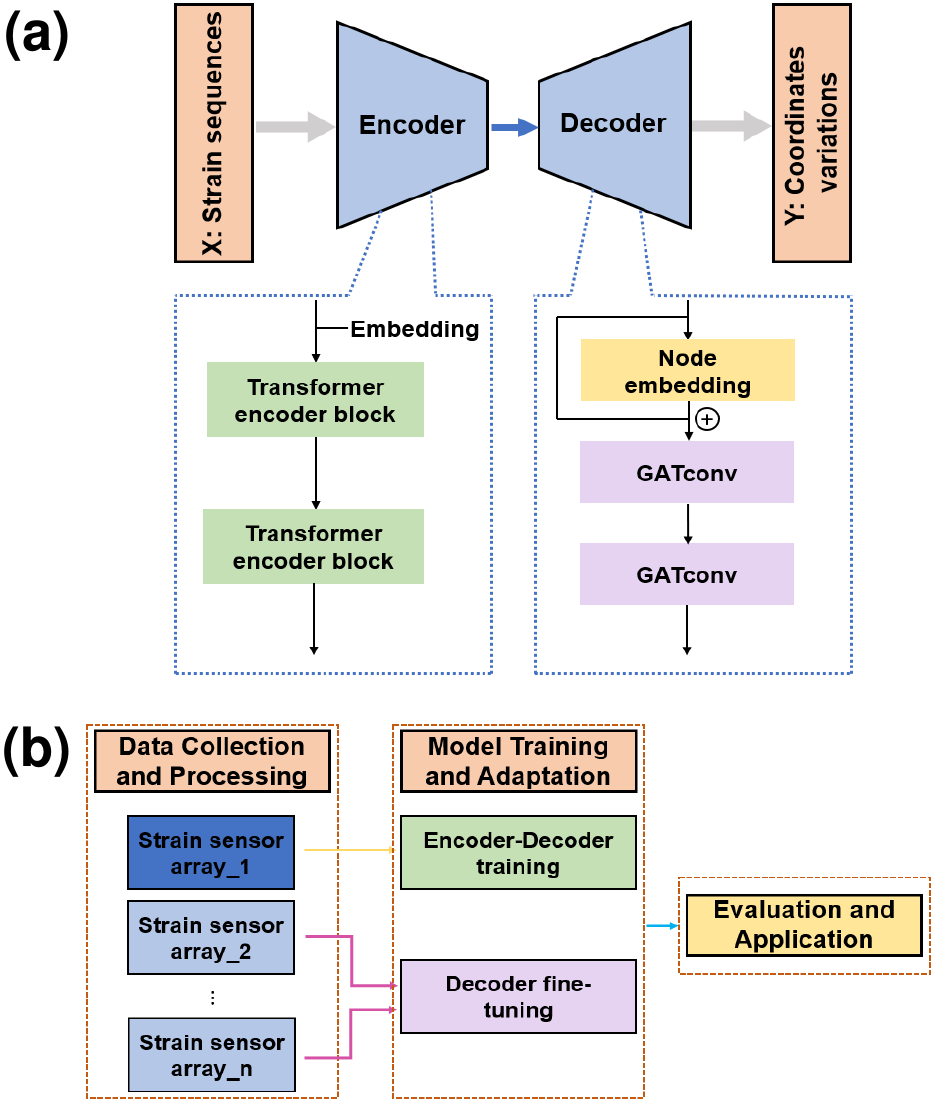}
\caption{\color{black}{\small Model structure and workflow of the proposed method. (a) Model structure. (b) Model training and cross-sensor adaptation workflow.}}
\label{model}
\vspace{-0.3cm}
\end{figure}

The whole structure and workflow can be visualized in \cref{model}. In practical applications, \textcolor{black}{only a few labeled samples collected with the stereo camera are needed for a totally new group of strain sensor array to fine-tune the model. Subsequently, the trained model is applied to surface shape sensing in various scenarios, using only strain sensor data.} The specific parameters have been described in Section \ref{model_parameters}.
\begin{algorithm}[H]
\caption{Meta-learning Training}
\label{algorithm}
\begin{algorithmic}[1]
        \Require 
        Labeled source data $S$ and labeled source target $T$
        \Ensure Average loss of tasks $L$ 
        
        \State Initialize virtual task number: n\_tasks and adaptation steps: n\_steps.
        
        \For{task = 1 \textbf{to} n\_tasks}:
            \State Virtual tasks generation: $virtual\_data$ and corresponding $label$.
            \State Clone model parameters: $clone\_parameters$.
            \For{step = 1 \textbf{to} n\_steps}:
                \State Compute loss with $virtual\_data$ and $label$.
                \State Update $clone\_parameters$.
            \EndFor
            \State Compute task loss: $task\_losses$.
            \State Update meta gradient: $meta\_grads$.
   
\EndFor
\State Update optimizer with average $meta\_grads$.
\State Clear the old gradients.
\end{algorithmic}
\end{algorithm}

\section{Experiments}
\label{sec:Experiments}

\subsection{Experimental Setup}

\begin{figure}[h]
\centering
\vspace{0.1cm}
\includegraphics[width=3.0in]{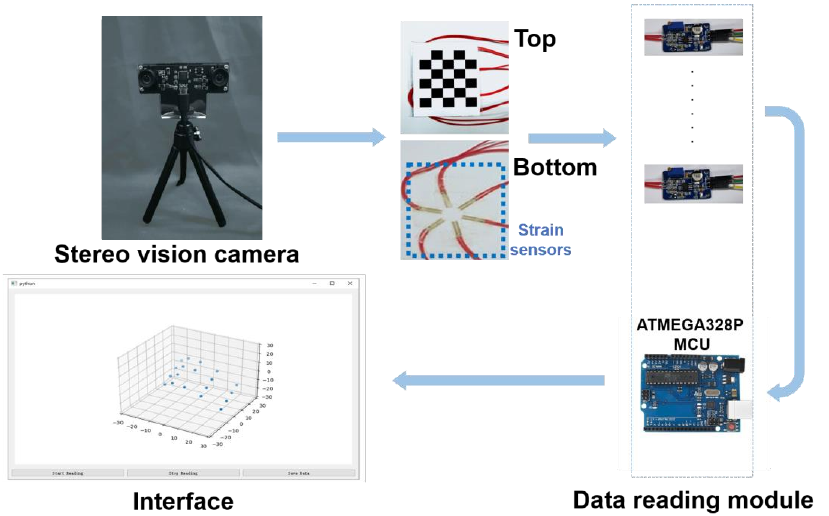}
\caption{\color{black}{\small Hardware system.}}
\label{Hardware system}
\vspace{-0.3cm}
\end{figure}

The hardware system consists of three main modules: a stereo camera, a PVC test surface \textcolor{black}{($65\ mm \times 53\ mm \times 0.6\ mm$)} with a \textcolor{black}{$4\times5$} chessboard test paper attached to \textcolor{black}{its} front surface and six strain sensors \textcolor{black}{mounted on its bottom surface around the center} at 60-degree intervals, and a data acquisition module. The stereo camera is a four-megapixel dual 1080P synchronized camera. The strain sensors, supplied by \textbf{Shenzhen RunesKee Technology Co., Ltd.}, use beryllium bronze as the sensitive material, with a detection area of 10.7 mm by 2.9 mm. Resistance drift relative to the average is 0.4 ohms, and resistance varies with bending direction, yielding increases or decreases. The sensitivity factor is 2 with a relative error within 1\%. The data acquisition module includes wheatstone bridge circuits and an arduino board with an \textbf{{ATMEGA328P}} MCU, communicating via serial at 250000 BPS for Python-based data capture. During data collection, strain data received via the serial port and the stereo camera video stream are recorded simultaneously, with timestamps aligned. \cref{Hardware system} illustrates the overall hardware system. Then, \textcolor{black}{four} different groups of strain sensors were tested using the system. A dataset was collected to train a surface shape sensing model of a sensor array, while the other three arrays utilize a substantially smaller volume of data for the adaptation process. The performances of these models were \textcolor{black}{later evaluated}.

\subsection{Data Synchronization between Stereo Camera and
Strain Sensor Array}
During data acquisition, strain sensors and stereo camera data were recorded synchronously with \textcolor{black}{different} acquisition rates, with the camera operating at 30 fps and the strain sensors at 8 fps. \textcolor{black}{Timestamps were recorded upon data arrival in both modalities}. The PVC surface was slowly deformed into various shapes while data \textcolor{black}{was} captured. The recorded video \textcolor{black}{was} then sampled at two-frame intervals, with each frame assigned a timestamp. Using these timestamps as references, a binary search \textcolor{black}{was} applied to find the closest strain sensor timestamp, linking strain values to the corresponding video frames. The synchronization error \textcolor{black}{was kept under 0.1 s, ensuring an accurate alignment between the strain measurements and the video data.}

\subsection{Test Processes}
Four sensor arrays designated as $a$, $b$, $c$, and $d$ are used for testing. \textcolor{black}{A large dataset from array $a$ is collected to train a high-performance model as the basic model, whereas, only small datasets from the other three arrays are used for adaptation. The detailed data volumes for each array are provided in Section \ref{results}}. Specifically, the array $b$ is used to visualize the shape sensing performances before and after adaptation, while the arrays $c$ and $d$ quantify error improvements \textcolor{black}{before and after adaptation. Finally, ablation experiments and analysis are conducted to validate the practical performance of each module and the proposed model. On the other hand, the origin of the feature points in all experiments was set to number 11, as its proximity to the geometric center of the shape results in more uniform and symmetric coordinate variations. Then, the adaptation processes below mainly contain the adjustment of the decoder learning rate and the proportion of frozen parameters.} Meanwhile, all experiments were conducted on a single NVIDIA A100 GPU.

\section{Results}
\label{results}
\subsection{Basic Model Performance of Sensor Array a}
The dataset acquired from the sensor array $a$ was used for basic model training, totally 12850 groups of data with a split ratio of 8:1:1 for training, validation, and test sets, respectively. The model was trained for 50 epochs using the Adam optimizer with a learning rate of 0.001. Meta-learning was performed every five batches, with 25 tasks (n\_tasks) and 15  steps (n\_steps) per meta-learning iteration. Upon completion of training, the model achieved a loss of 2.79 in this sensor array and a mean absolute distance error of 2.21 mm in its test set. 

\subsection{Performance Visualization of Sensor Array $b$}
\label{new_sensor}
\begin{figure*}[h]
\centering
\vspace{-0.3cm}
\includegraphics[width=6.8in]{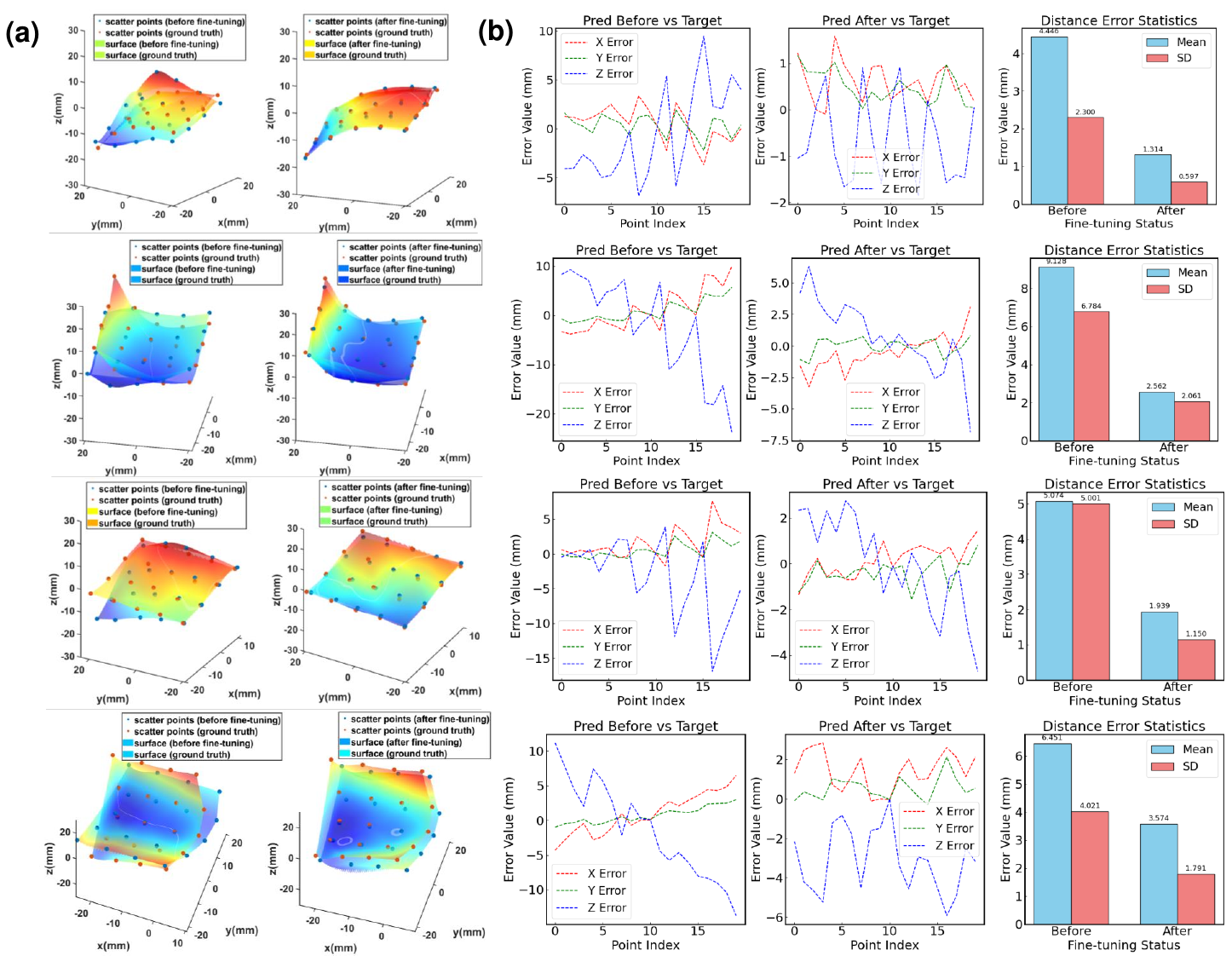}
\caption{\textcolor{black}{\small Typical surface sensing results of sensor array $b$. (a) Shape sensing results. (b) Error comparisons of results: \textbf{Pred Before} means calculating before adaptation, \textbf{Pred After} means calculating after adaptation and distance error statistics (Mean and Standard Deviation).}}
\label{shape}
\vspace{-0.3cm}
\end{figure*}

\textcolor{black}{For the sensor array $b$ in this part, the recorded data was about half that of the array $a$}. Subsequently, we fine-tuned the decoder of the basic model using newly recorded data from this array. The new data \textcolor{black}{were} divided into 8:1:1 as training set, validation set, and test set. \textcolor{black}{When the decoder was fine-tuned, the training set was used and the adaptation steps, learning rate, unfreeze parameter ratio, and training set ratio values} were set as 20, 0.003, 0.3, and 0.1, respectively. That is, 510 groups of new data that are less than $5.0\%$ of the sensor array $a$ were used for adaptation. The average adaptation time of the sensor array $b$ was less than 1 s. Then the typical surface sensing results in test set of the sensor array $b$ with the adapted model can be visualized in \cref{shape}. In \cref{shape} (a), four different shapes are visualized, and it can be noticed that without adaptation, \textcolor{black}{the sensing results do not} accurately reflect the characteristics of these shapes. After adaptation, the model improves a lot and can be used for different shapes. In \cref{shape} (b), the distance errors of the points in these four shapes are illustrated, and most errors arise along the Z axis. Compared to the results without adaptation, the results after adaptation demonstrate significant improvement across all three axes. \textcolor{black}{For each shape, the maximum values of $|C_{x(i,j)}|$ and $|C_{y(i,j)}|$ in \eqref{curvature} were computed, and the larger of the two ($|C_{x(i,j)}|$ or $|C_{y(i,j)}|$) was used as the metric for analysis. The performances were improved from 45.90, 42.35, 44.31 and 42.88 mm to 40.97, 42.65, 38.85 and 42.90 mm, respectively, while the corresponding ground-truth values are 42.58, 45.10, 39.17 and 43.05 mm}. Then 500 samples were tested to evaluate the performance of the adapted model. The mean distance error across all points was 3.30 mm—only 1.09 mm higher than \textcolor{black}{sensor the} array a. \textcolor{black}{After fine-tuning, the accuracy improved by 47.2\%, 39.4\%, and 47.6\% along the $X$, $Y$, and $Z$ axes, respectively}. By contrast, using the model without adaptation on this \textcolor{black}{sensor array yielded} a mean distance error of 6.25 mm, causing significant deviations in the sensed shape.

\begin{figure}[h]
\centering
\vspace{0cm}\includegraphics[width=3.4in]{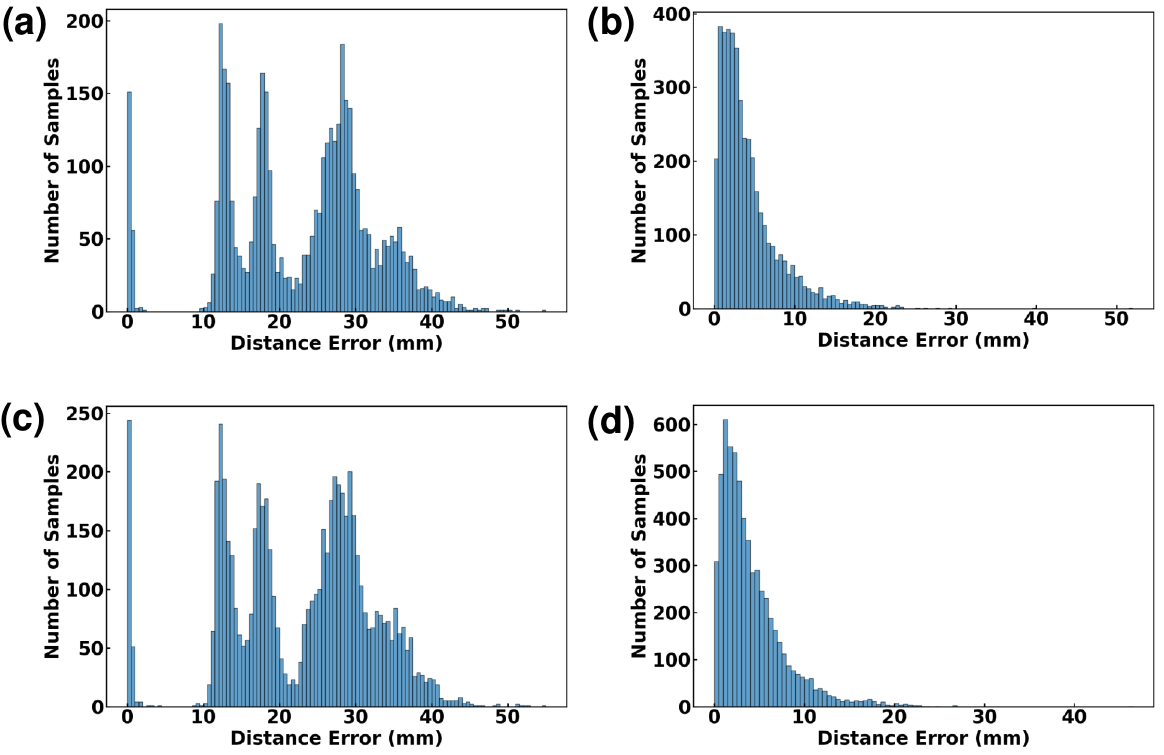}
\caption{\small Evaluation of two other sensor arrays $c$ and $d$. (a) Results of sensor array $c$ without adaptation. (b) Results of sensor array $c$ with adaptation. (c) Results of sensor array $d$ without adaptation. (d) Results of sensor array $d$ with adaptation.}
\label{sensors}
\vspace{-0.3cm}
\end{figure}

\begin{figure}[hp]
\centering
\vspace{0.1cm}
\includegraphics[width=3.5in]{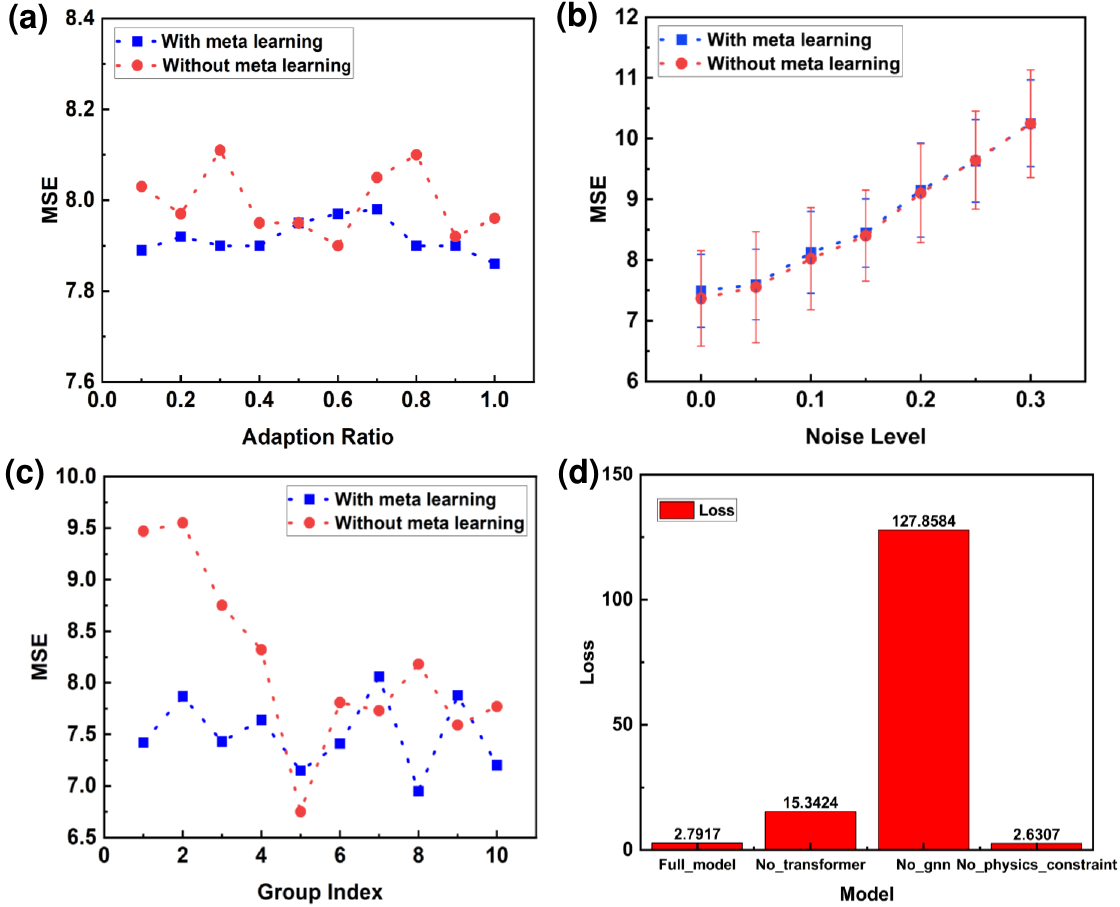}

\caption{\small Evaluation with different adapted parameters and without different modules. (a) Performance of different adaptation ratios of data from a new sensor array. (b) Performance of different noise levels of data from a new sensor array. (c) Performance of ten groups of test data from a new sensor array. (d) Training performance without different modules.}
\label{ablation}
\vspace{-0.3cm}
\end{figure}

\begin{figure}[h]
\centering
\vspace{0cm}
\includegraphics[width=3.4in]{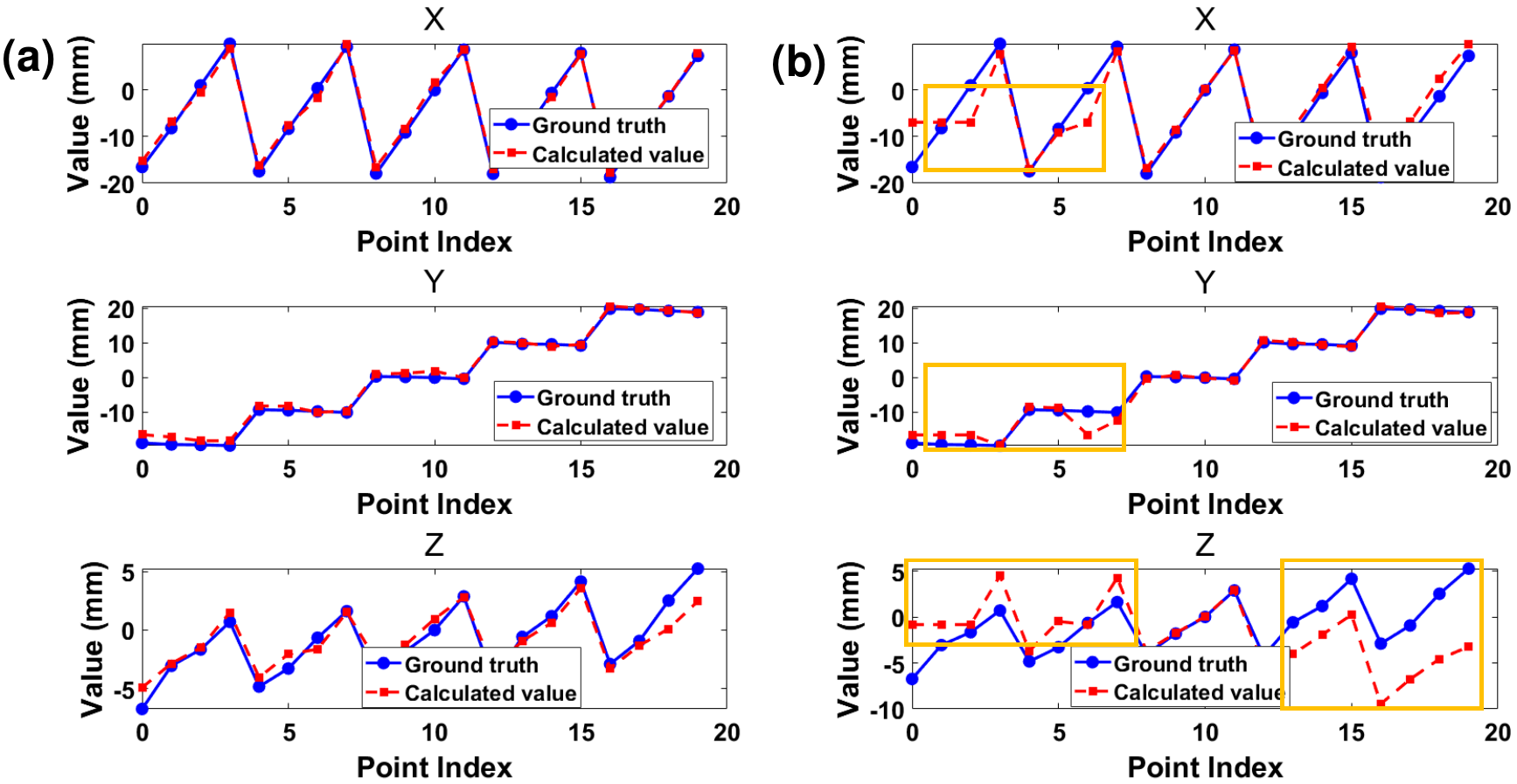}
\caption{\textcolor{black}{\small Results of coordinates with/without constraint. (a) With constraint. (b) Without constraint.}}
\label{training}
\vspace{-0.3cm}
\end{figure}

\subsection{Evaluations with Strain Sensor Array $c$ and $d$}
  In this part, for the other two sensor arrays $c$ and $d$, the data collection and processing procedures were consistent with those described in Section \ref{sec:Experiments}. The data \textcolor{black}{volumes} used for adaptation were 170 and 245 for these two arrays and the test results can be seen in \cref{sensors}. It can be noticed that the adapted models (\cref{sensors} (b) and \cref{sensors} (d)) both \textcolor{black}{performed better} than those without adaptations (\cref{sensors} (a) and \cref{sensors} (c)). Meanwhile, the model trained with meta-learning showed better performance after adaptation compared to the model without meta-learning. The errors of these two sensor arrays improved from 4.34 mm to 4.05 mm and from 4.67 mm to 4.10 mm compared to the models without meta-learning, respectively. Without adaptation, the errors were about 23.0 mm for these two sensor arrays. Finally, the statistics showed that\textcolor{black}{,} after adaptation, the number of samples with errors within 5.0 mm increased from 5.0\% to 71.0\% and 70.5\% for these two sensor arrays, respectively.

\subsection{Ablation Experiment and Analysis}
Different \textcolor{black}{parameter settings have been used} to evaluate the performance of the proposed method under different conditions, and \cref{ablation} visualizes the results. \cref{ablation} (a) shows the results in the test set of sensor array $b$ for 100 rounds. It can be noticed that the proposed encoder-decoder architecture itself exhibits strong post-fine-tuning generalization performance. However, \textcolor{black}{compared with the model without meta-learning, introducing meta-learning reduces the sensitivity of the model to the proportion of new data, leading to more stable performance and smaller error fluctuations.} This observation is further supported by the results in \cref{ablation} (b). Random noise \textcolor{black}{ with an amplitude ranging from 0\% to 30\% of the original data magnitude was added, and} 100 test sets were evaluated at each noise level. The error curve increases with higher noise levels, and the average MSEs of models with or without meta-learning remain close. However, in all results, the models with meta-learning exhibit consistently narrower standard deviation ranges compared to those without meta-learning, which indicates that the model with meta-learning has achieved an improved level of noise robustness. As shown in \cref{ablation} (c), results from 10 groups of \textcolor{black}{randomly selected} data in the test set indicate that, without meta-learning, the model exhibits unstable adaptability to new samples, with significant performance fluctuations. In contrast, meta-learning yields more consistent results, \textcolor{black}{corroborating previous analysis}. \textcolor{black}{In addition}, the average MSEs in \cref{ablation} (c) are 7.57 and 8.07, respectively. The results in \cref{ablation} (d) demonstrate the significant contributions of the Transformer and graph neural network components to the proposed model. \textcolor{black}{In \cref{training}, an example illustrating the effect of the constraint is presented. It can be clearly seen that, when the constraint is removed from the loss function, the calculated coordinates may exhibit abrupt variations and even extremely large errors, as highlighted by the orange box in the figures. This demonstrates that the introduced constraint plays a crucial role in the model.}

\subsection{Comparisons with Other Work}
Many other methods have been used for surface shape sensing, and the specific performance of each method can be visualized in \cref{tab: different methods}. It is very difficult to achieve high shape sensing accuracy while the device is easily portable. \textcolor{black}{On the other hand, all these methods that combine different sensor data with machine learning or deep learning fail to consider cross-sensor generalization.} In real applications, there are significant differences in characteristics even among sensors of the same type. Meanwhile, it is impractical to \textcolor{black}{perform labor-intensive calibration and data collection work repeatedly to adapt sensors to their corresponding applications}. This paper is also the first to propose a potential solution to this problem from this perspective, achieving favorable results.

\begin{table*}
\centering
\vspace{0.1cm}
\caption{Comparison of different methodologies}
\label{tab: different methods}
\setlength\tabcolsep{8pt}
\begin{tabular*}{1\linewidth}{c c c c c c }
\toprule
 &  \multirow{3}*{\makecell{Techniques}} & \multirow{3}*{\makecell{Easy to integrate \\and portable}} &\multirow{3}*{\makecell{Shape sensing\\ dimension}} & \multirow{3}*{\makecell{ Average accuracy of \\shape sensing}} & \multirow{3}*{\makecell{Cross-sensor \\adaptation}}\\
 & & & & &\\   
 & & & & &\\
\midrule
\cite{4} & Capacitive sensors + Optical tracking + Deep learning & Yes & 3 & \textless 5 mm & No\\
\cite{8} & FBG + Optical tracking + Deep learning
 & No & 3 & \textless 2 mm & No\\
\cite{9} & Strain sensors + Geometric computation
 & Yes & 2 & 7.15 mm & /\\
\cite{10} & Piezoelectric sensors + Vision + 
Linear regression
 & Yes & 3 & \textgreater 10 mm & No\\
\cite{12} & Capacitive sensors + Geometric computation & Yes & 2 & Relevant to radii & / \\
This Paper & Strain sensors + Vision + Deep learning & Yes & 3 & $\sim 4.0$ mm & Yes\\

\bottomrule
\end{tabular*}
\end{table*}

\section{Discussion and Limitation}
\subsection{Discussion}
This paper proposes a robust surface shape sensing method with cross-sensor adaptation ability. Given the challenge of few-shot adaptation across different sensor arrays, \textcolor{black}{an encoder-decoder architecture and training it through meta-learning are proposed}—ultimately developing a cross-sensor adaptation solution that enables stable shape sensing. Different groups of sensors may experience slight displacement because of installation when attached to the measured surface, which further demonstrates that the algorithm also has strong advantages in interference resistance.
Experimental results show that this method achieves high accuracy in surface shape sensing. Furthermore, the developed system is not only simple in structure but also portable, allowing easy integration in other scenarios that rely on sensing technology, such as flexible robotics and flexible electronic skins.
\subsection{Limitations and Potential Improvements}
There is still room for improvement in the future. \textcolor{black}{Firstly, although data collected for training covered many bending types and degrees such as bending in the longitudinal and transverse directions, single-corner, multiple-corner, and diagonal bending and so on, it was not enough. Thus, the method’s performance under more complex deformations remains untested. However, the results suggest the method is stable and has great potential to perform well in such cases. Secondly, the number of sensing elements in the array, as well as their spatial arrangement, should also be investigated to optimize performance for more complex shapes.} Then, finite element simulation can incorporate data of more virtual feature points based on physical knowledge and existing feature points, and combining this with deep learning methods could potentially achieve better results. Finally, unsupervised learning shows great future potential, especially if cross-sensor adaptation can be done without extra labeled data, which would greatly increase applicability and impact.

\section{Conclusion}
This research proposed a transformer encoder and graph neural network decoder architecture with meta-learning, achieving robust surface shape sensing and excellent cross-sensor adaptation \textcolor{black}{performance}. The proposed method can be adapted to a new sensor array with less than $5.0\%$ new labeled data in less than 1 second with an average error of approximately 4.0 mm. The proposed method holds promise in fields such as soft electronic skin, soft robotics and wearable devices.

\bibliographystyle{unsrt}
\bibliography{ref}
\end{document}